\def\year{2017}\relax
\documentclass[letterpaper]{article} 
\usepackage{aaai18}  
\usepackage{natbib}
\usepackage{times}  
\usepackage{helvet}  
\usepackage{courier}  
\usepackage{url}  
\usepackage{graphicx}  
\usepackage{subcaption}
\usepackage{amsmath}
\usepackage{amsfonts}
\usepackage{color}
\begin{document}
%
\title{A Computational Model of 
Commonsense Moral Decision Making}

\author{Richard Kim \And Max Kleiman-Weiner \And Andr\'es Abeliuk \AND 
Edmond Awad \And Sohan Dsouza \And Josh Tenenbaum \And Iyad Rahwan \AND \textmd{Massachusetts Institute of Technology, Cambridge MA, USA}
}

\maketitle

%
%
\begin{abstract}
We introduce a new computational model of moral decision making, drawing on a recent theory of commonsense moral learning via social dynamics. Our model describes moral dilemmas as utility function that computes trade-offs in values over abstract moral dimensions, which provide interpretable parameter values when implemented in machine-led ethical decision-making. Moreover, characterizing the social structures of individuals and groups as a hierarchical Bayesian model, we show that a useful description of an individual's moral values -- as well as a group's shared values -- can be inferred from a limited amount of observed data. Finally, we apply and evaluate our approach to data from the Moral Machine, a web application that collects human judgments on moral dilemmas involving autonomous vehicles.
\end{abstract}

%
%
Recent advances in machine learning, notably Deep Learning, have demonstrated impressive results in various domains of human intelligence, such as computer vision \citep{Szegedy}, machine translation \citep{Wu2016}, and speech generation \citep{Oord2016}. In domains as abstract as human emotion, Deep Learning has shown a proficient capacity to detect human emotions in natural language text \citep{Felbo2017}. These achievements indicate that Deep Learning will be paving the way for AI in ethical decision making. 

However, training Deep Learning models often requires human-labeled data numbering in the millions, and despite recent advances that enables a model to be trained from a small number of examples \citep{Vinyals2016MatchingLearning,Santoro2016One-shotNetworks}, this constraint remains a key challenge in Deep Learning. In addition, Deep Learning models have been criticized as ``blackbox" algorithms that defy attempts at interpretation \citep{Lei2016RationalizingPredictions}. The viability of many Deep Learning algorithms for real-world applications in business and government has come into question as a recent legislation in the EU, slated to take effect in 2018, will ban automated decisions, including those derived from machine learning if they cause an ``adverse legal effect" on the persons concerned \citep{Goodman2016EuropeanExplanationquot}.

In contrast to Deep Learning algorithms, evidence from studies in human cognition suggests that humans are able to learn and make predictions from a much smaller number of noisy and sparse examples \citep{Tenenbaum2011}.  Moreover, studies have shown that humans are able to internally rationalize their moral decisions and articulate reasons for these \citep{Haidt2001TheJudgment.}. Given this stark difference between the current state of machine learning and human cognition, how can we draw on the latest theories in cognitive science to design AI with the capacity to learn moral values from limited interactions with humans and make decisions with explicable processes?

A recent theory from the field of cognitive science postulates that humans learn to make ethical decisions by acquiring abstract moral principles through observation and interaction with other humans in their environment \citep{Kleiman-Weiner2017}. This theory characterizes ethical decision as utility maximizing choice over a set of outcomes whose values are computed from weights people place on abstract moral concepts such as ``kin" or ``reciprocal relationship." In addition, given the dynamics of individuals and their memberships in a group, the framework explains how an individual's moral preferences, and the actions resulting from them, lead to a development of the group's shared moral principles (i.e. group norms).        

In this work we extend the framework introduced by \cite{Kleiman-Weiner2017} to explore a computational model of the human mind in moral dilemmas with binary-decisions. We characterize the decision making in moral dilemmas as a utility function that computes the trade-offs of values perceived by humans in the choices of the dilemma. These values are the weights that humans put on abstract dimensions of the dilemma; we call these weights {\em moral principles}. Furthermore, we represent an individual agent as a member of a group with many other agents that share similar moral principles; these shared moral principles of the group as an aggregate give rise to the {\em group norm}. Exploiting the hierarchical structure of individuals and group, we show how hierarchical Bayesian inference \citep{Gelman2013} can provide a powerful mechanism to rapidly infer individual moral principles as well as the group norm with sparse and noisy data.

We apply our model to the domain of autonomous vehicles (AV) through a data set from the Moral Machine, a web application that collects human judgments in ethical dilemmas involving AV.\footnote{http://moralmachine.mit.edu/} A recent study on public sentiment on AV reveals that endowing AI with human moral values is an important step before AV can undergo widespread market adoption \citep{Bonnefon2016}. In light of this study, we view application of our model to understand how the human mind perceives and resolves moral dilemmas on the road as an important step towards building an AV with human moral values.

This paper makes the following distinct contributions towards building an ethical AI:
\begin{itemize}
\item Introducing a novel computational model of moral decision making that characterizes moral dilemma as a trade-off of values along abstract moral dimensions.  We show that this model well-describes how the human mind processes moral dilemmas and provides an interpretable process for an AI agent to arrive at a decision in a moral dilemma.
\item Characterizing the social structure of individuals and groups as a hierarchical Bayesian model, we show that the model can rapidly infer moral principles of individuals from limited number of observational data.  Rapidly inferring other agents' unique moral values will be crucial, as AI agents interact with other agents, including humans.   
\item Demonstrating the model's capacity to rapidly infer group's norms, characterized as prior over individual moral preferences. Inferring shared moral values of a group is an important step towards designing an AI agent that makes socially optimal choices.
\end{itemize}

%
%
\section{Moral Machine Data}
Moral Machine is a web application built to collect and analyze human perceptions of moral dilemmas involving autonomous vehicles. As of October 2017, the application has collected over $30$ million responses from over $3$ million unique respondents from over $180$ countries around the world. Here, we briefly describe the design of moral dilemma and data structure in Moral Machine.

\begin{figure}[h]
    \centering
    \includegraphics[width=0.35\textwidth]{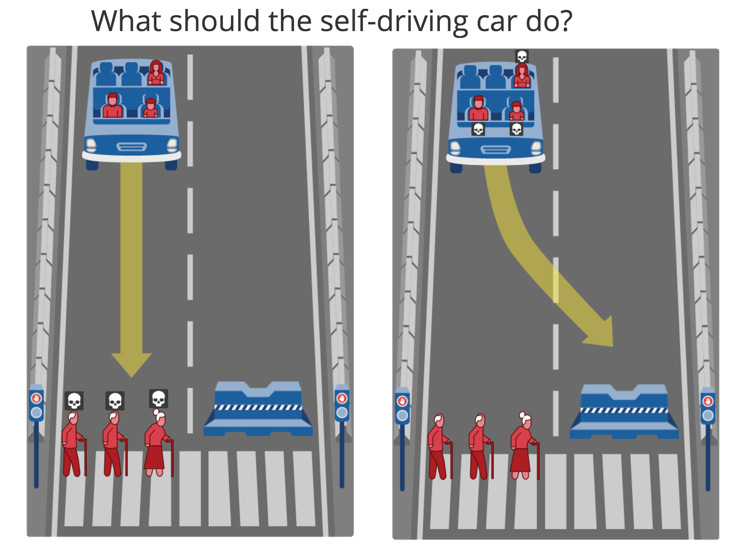}
    \caption{{\itshape Moral Machine} interface. An example of a moral dilemma that features an AV with sudden brake failure, facing a choice between either not changing course, resulting in the death of three elderly pedestrians crossing on a ``do not cross'' signal, or deliberately swerving, resulting in the death of three passengers; a child and two adults.}
    \label{fig:moral_machine}
\end{figure}

\begin{figure}[h]
    \centering
    \includegraphics[width=0.35\textwidth]{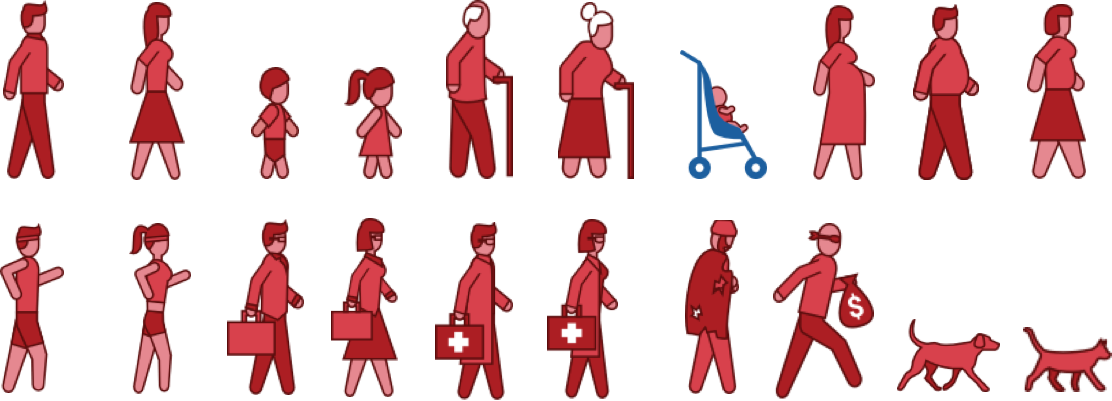}
    \caption{Twenty characters in Moral Machine represent various demographic attributes such as gender, age, social status, fitness level, and species.}
    \label{fig:moral_machine_characters}
\end{figure}

In a typical Moral Machine session, a respondent is shown $13$ scenarios such as the example shown in Figure \ref{fig:moral_machine}. In each scenario, the respondent is asked to choose one of two outcomes that have different ethical consequences with different trade-offs. A scenario can contain any random combination of twenty characters (see Figure \ref{fig:moral_machine_characters}) that represents various demographic attributes found in a general population. In addition to the demographic factors, Moral Machine scenario also includes the factors of character's status as a passenger or a pedestrian and its status as a pedestrian who is crossing on green light or red light.         

In addition to the respondents' decisions, data about their response duration (in seconds) to each scenario and their approximate geo-location is also collected. This allows us to infer the country or region of access.

Every scenario has two choices, which we represent as a random variable $Y$ with two realizable values $\{0, 1\}$. A respondent's choice to swerve (i.e., intervene) is represented as $Y=1$, and likewise, their choice to stay (i.e., not-intervene) is represented as $Y=0$. The respondent's choice yields a state of certain set of characters being saved over others. The resultant state is represented by character vector $\Theta_y \in \mathbb{N}^{K}$, which denotes the resultant state of choice $y$. 

\begin{figure}[h]
    \centering
    \includegraphics[width=0.25\textwidth]{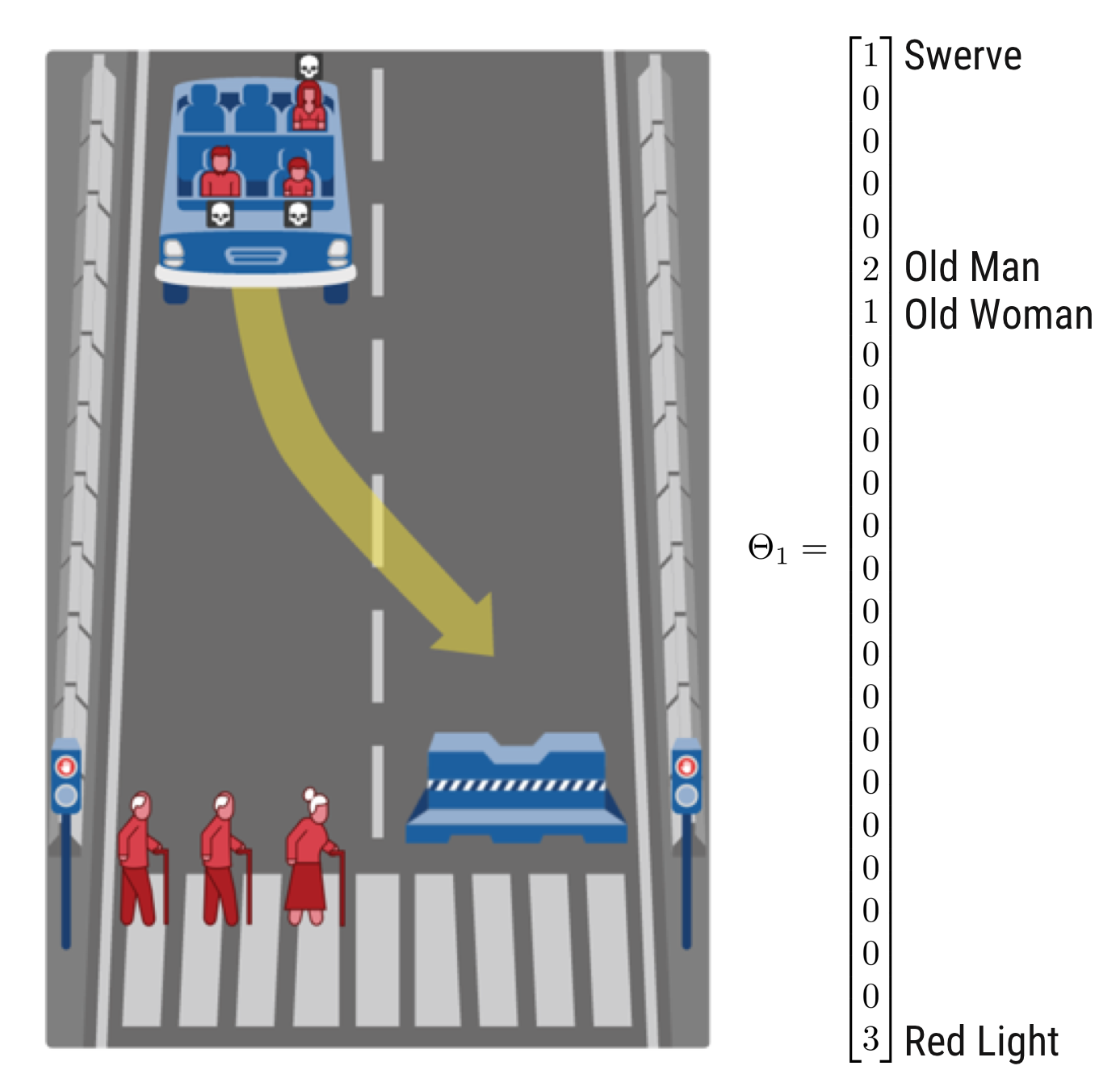}
    \caption{An example of vector representation of a state in the Moral Machine character space.}
    \label{fig:example1}
\end{figure}

As an illustration, we show a vector representation of a resultant state of swerve in Figure \ref{fig:example1}.  The vector element of {\em old man} character is denoted by value of $2$, representing two {\em old man} characters that will be saved from the choice of swerve $Y=1$. In addition, the vector element of {\em red light} feature is denoted by value of $3$, representing three pedestrian who are crossing the red light.

%
%
\section{Moral Dilemma as Utility Function}
Jeremy Bentham, the founder of modern utilitarian ethics, described ethical decision in a moral dilemma as a utility maximizing decision over the sum of trade-offs over values in the dilemma \citep{Bentham1789AnLegislation}. More recently, cognitive psychologists have formalized the idea of analyzing moral dilemma using utility function that computes various  trade-offs in the dilemma \citep{Mikhail2007, Mikhail2011ElementsCognition}. Evidence of moral decision making in young children suggests that children base their moral judgments by computing trade-off of values over abstract concepts \citep{Kohlberg1981EssaysDevelopment}.

Using this framework, we can analyze how a respondent arrives to his/her decision based on the values that he/she places on abstract dimensions of the moral dilemma, which we label {\em moral principles}. For instance, when a respondent chooses to save a female doctor character in a scenario over an adult male character, this decision is in part due to the value that respondent places on the abstract concept of {\em doctor}, a rare and valuable member in society who contributes to improvement of social welfare. The abstract concept of {\em female} gender also would be a factor in his or her decision.  

\begin{figure}[h]
    \centering
    \includegraphics[width=0.4\textwidth]{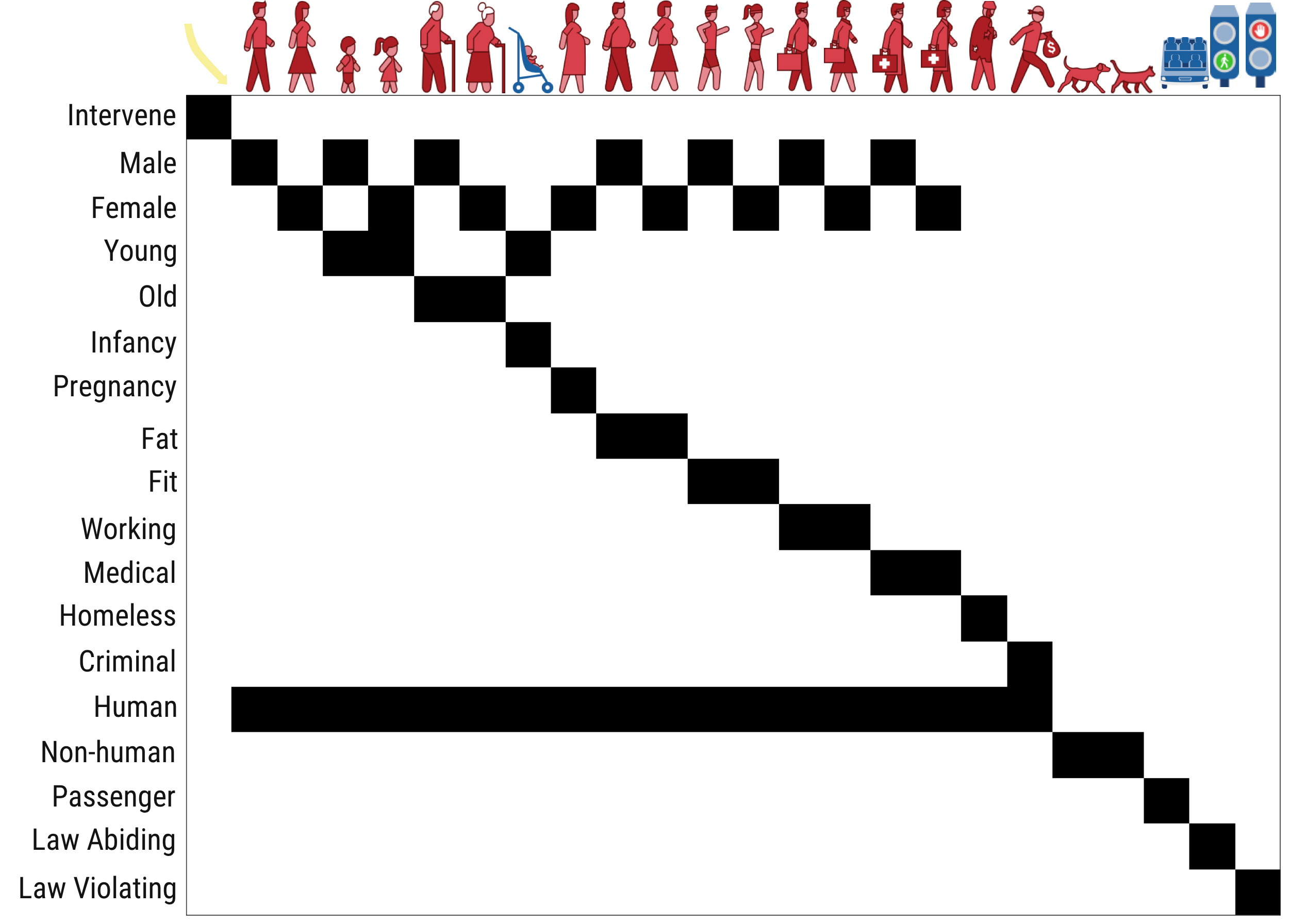}
    \caption{An example of a binary matrix $A$ that decomposes  the characters in Moral Machine into abstract features.  Black squares indicate the presence of abstract features in the characters.}
    \label{fig:feature_matrix}
\end{figure}

In Moral Machine, twenty characters share many abstract features such as {\em female}, {\em elderly}, {\em non-human}, etc.  Hence, the original character vector $\Theta_y$ can be decomposed into a new vector in the abstract feature space $\Lambda_y \in \mathbb{N}^{D}$ where $D \leq K$ via feature mapping $F : \Theta \rightarrow \Lambda$.  In this work, we use a linear mapping $F(\Theta) = A\Theta$ where $A$ is a $18 \times 24$ binary matrix such as the one shown in Figure \ref{fig:feature_matrix}. 

\begin{figure}[h]
    \centering
    \includegraphics[width=0.32\textwidth]{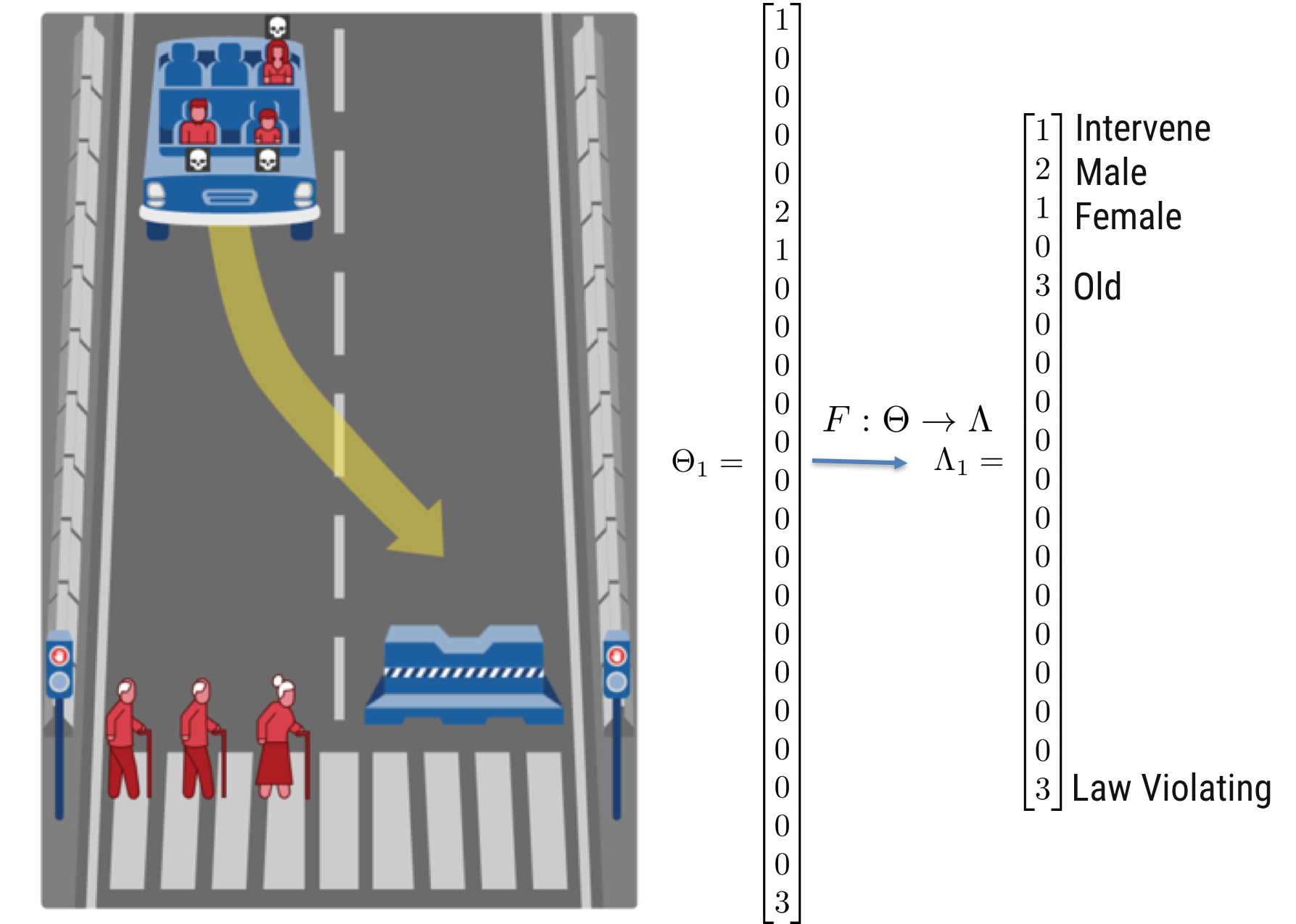}
    \caption{Vector representation of abstract features of a scenario choice.}
    \label{fig:example2}
\end{figure}

Shown in Figure \ref{fig:example2}, the original state vector in the Moral Machine character space $\Theta$ is mapped into a new state vector in the abstract feature space $\Lambda$.  We note that vector element of {\em old} is denoted by value of $3$ representing three character with this feature.    

We define moral principles as weights $w \in \mathbb{R}^{D}$ that respondent place along the $D$ abstract dimensions $\Lambda$. These weights represent how the respondent values abstract features such as {\em young}, {\em old}, or {\em doctor} to compute utility value of their choices. For simplicity, we model the utility value of a state as a linear combination of the features in the abstract dimension:
\begin{equation}
u(\Theta_{i}) = w^{\top}F(\Theta_{i})
\end{equation}

With utility values of the choice to not-intervene $u(\Theta_0)$ and intervene $u(\Theta_1)$, respondent's decision to intervene $Y=1$ is seen as probabilistic outcome based on sigmoid function of net utility of the two choices:
\begin{equation}\label{eq:sigmoid}
P(Y=1|\Theta) = \frac{1}{1+e^{-U(\Theta)}}
\end{equation}
where
\begin{equation}\label{eq:netutility}
U(\Theta) = u(\Theta_1) - u(\Theta_0).
\end{equation}

We turn our attention to inferring individual moral principles of respondents from sparse and noisy observation of their decisions in moral dilemma.

%
%
\section{Hierarchical Moral Principles}
Studies by anthropologists have shown that societies across different regions and time periods hold widely divergent views about what actions are ethical \citep{10.2307/2677736, House2013, Blake2015}. For example, certain societies  strongly emphasize respect for the elderly while others focus on protecting the young.  These views in a society are what we refer to as the society's {\em group norms}.  

Nevertheless, even in a society with a homogeneous cultural and ethnic make-up, individual members of the group can hold unique and different moral standards \citep{Graham2009LiberalsFoundations.}. How can we model the complex relationship between the group norm and individual moral principles?

We introduce {\em hierarchical moral principles} model, which is an instance of hierarchical Bayesian model \citep{Gelman2013}. Returning to data in Moral Machine, consider $N$ respondents that belong to a group $g \in G$. This group can be a country, a culture, or a region within which customs and norms are shared.  

The moral principles of respondent $i$ is drawn from a multivariate normal distribution parameterized by the mean values of the group $w^g$ on the $D$ dimensions:

\begin{equation}
w_{i} \sim Normal_{D}(w^{g}, \Sigma^g),
\end{equation}

\noindent where the diagonal of the covariance matrix $\Sigma^g$ represents the in-group variance or differences between the members of the group along the abstract dimensions. Higher variance value describes broader diversity of opinions along that corresponding abstract dimension. In addition, covariance (off-diagonal) values capture the strength of relationship between the values they place on  abstraction dimension. As an example, a culture that highly values {\em infancy} should also highly value {\em pregnancy} as they are intuitively closely related concepts. Covariance matrix allows the Bayesian learner to understand related concepts and use the relationship to rapidly approximate the values of one dimension after inferring that of a highly correlated dimension.

\begin{figure}[h]
    \centering
    \includegraphics[width=0.3\textwidth]{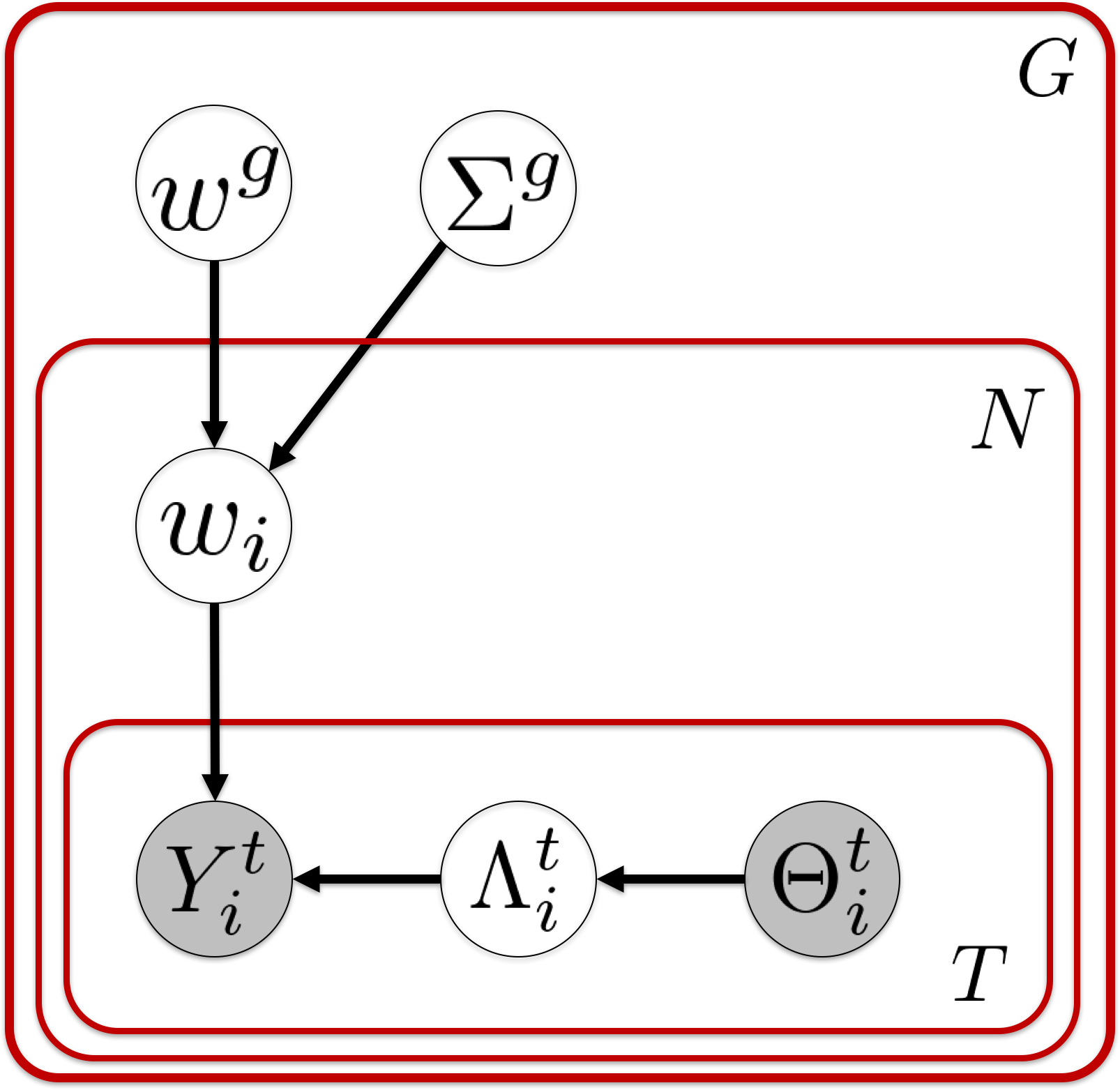}
    \caption{Graphical representation of hierarchical Bayesian model of moral principles.}
    \label{fig:graphical_model}
\end{figure}

Let ${\bf w} = \{w_{1}, ..., w_{i}, ... ,w_{N}\}$ be a set of unique moral principles by $N$ respondents.  Each respondent $i$ makes judgments on $T$ scenarios ${\bf \Theta} = \{\Theta^{1}_{1},...,\Theta^{t}_{i},..., \Theta^{T}_{N}\}$. Judgment by respondent $i$ is an instance of a random variable $Y^{t}_{i}$. Given the observation of the set of states ${\bf \Theta}$ and the decisions ${\bf Y}$, the posterior distribution over the set of moral principles follows:
\begin{equation}
\begin{split}
P({\bf w}, w^{g}, \Sigma^{g}|{\bf \Theta}, {\bf Y}) \propto  P({\bf \Theta}, & {\bf Y}|{\bf w})P({\bf w}|w^{g}, \Sigma^g) \\
& P(w^{g})P(\Sigma^g)
\end{split}
\end{equation}
where the likelihood is
\begin{equation}
P({\bf \Theta},{\bf Y}| {\bf w}) = \prod_{i=1}^{N}\prod_{t=1}^{T} p_{ti}^{y^{t}_{i}}(1-p_{ti})^{(1-y^{t}_{i})}
\end{equation}
and $p_{ti} = P(Y^{t}_{i} = 1| \Theta^{t})$ is the probability that a respondent chooses to swerve in scenario $t$ given $\Theta^{t}$ as shown in Equation \ref{eq:sigmoid}. Graphical representation of the model is presented in Figure \ref{fig:graphical_model}.

As an illustration, we randomly sampled 99 respondents from Denmark, which equates to 1,287 response data. We specified prior over the covariance matrix $P(\Sigma^g)$ with LKJ covariance matrix \citep{Lewandowski2009} with parameter $\eta=2$ :
\begin{equation}
\Sigma^g \sim LKJ(\eta)
\end{equation}
and the prior over group weights $P(w^g)$ with 
\begin{equation}
w^g \sim Normal_{D}(\mu, \Sigma^g)
\end{equation}
where $\mu = {\bf 0}$.

\begin{figure*}[h]
    \centering
    \begin{subfigure}[b]{0.4\textwidth}
        \includegraphics[width=\linewidth]{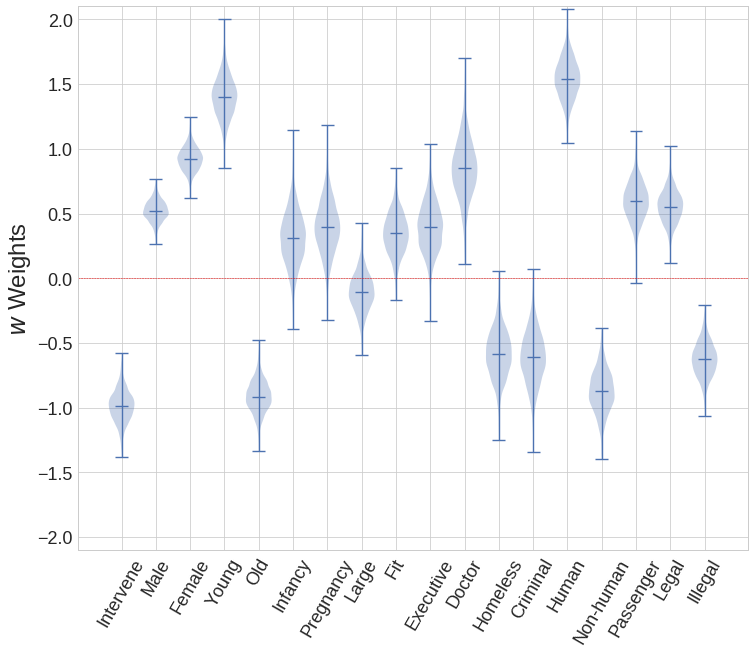}
        \caption{}
        \label{fig:denmark_weights}
    \end{subfigure}
    \begin{subfigure}[b]{0.4\textwidth}
        \includegraphics[width=\linewidth]{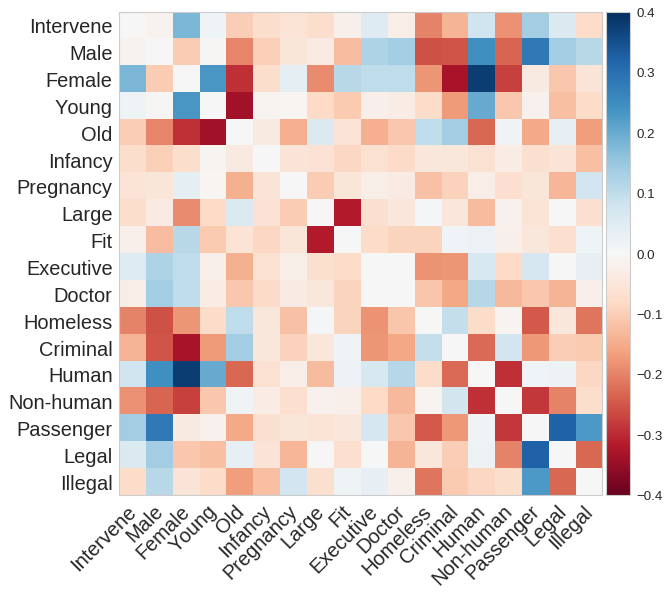}
        \caption{}
        \label{fig:denmark_correlation}
    \end{subfigure}
    \begin{subfigure}[b]{0.33\textwidth} 
        \includegraphics[width=\linewidth]{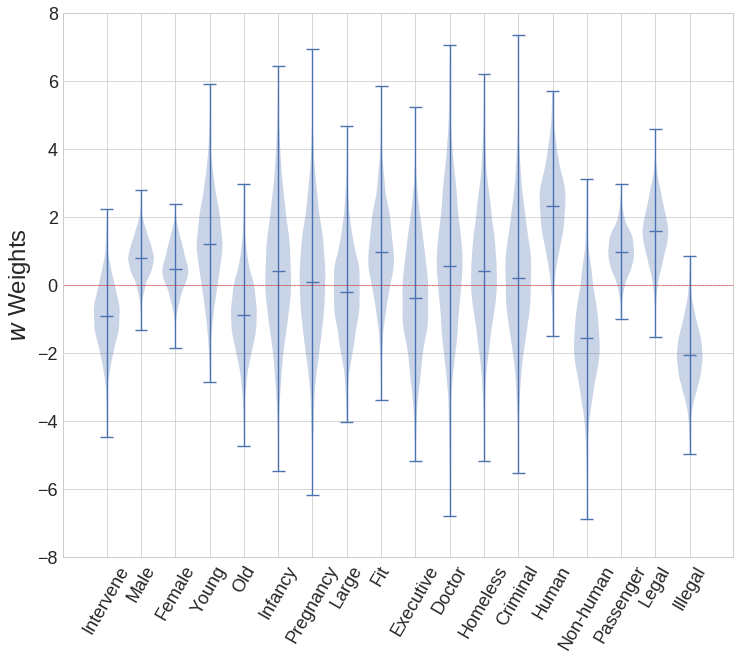}
        \caption{}
        \label{fig:denmark_weights_1}
    \end{subfigure} \hfill
    \begin{subfigure}[b]{0.33\textwidth} 
        \includegraphics[width=\linewidth]{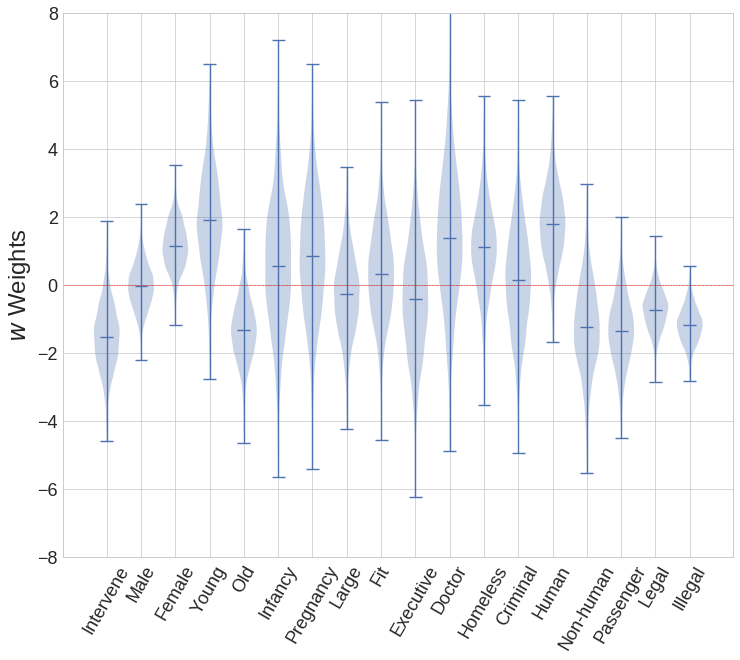}
        \caption{}
        \label{fig:denmark_weights_2}
    \end{subfigure} \hfill
    \begin{subfigure}[b]{0.33\textwidth} 
        \includegraphics[width=\linewidth]{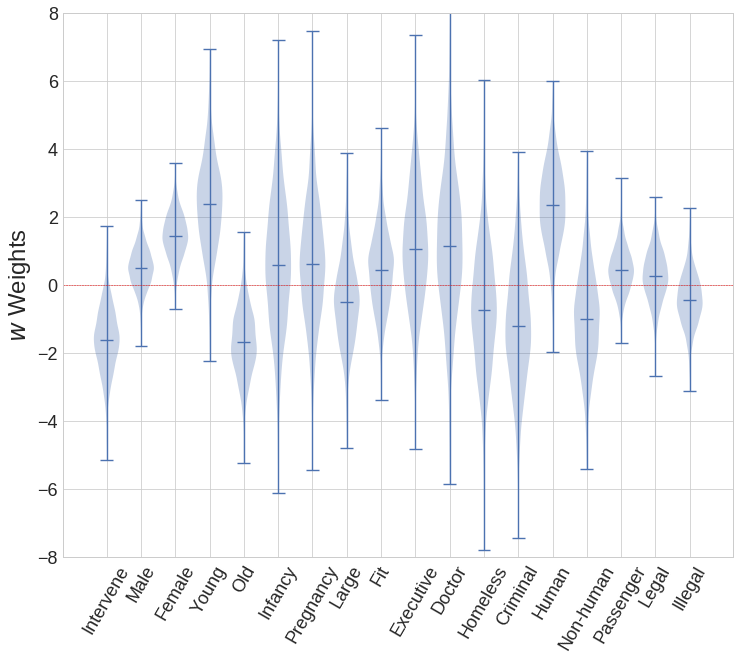}
        \caption{}
        \label{fig:denmark_weights_3}
    \end{subfigure}
    \caption{(a) Inferred group norm of sampled Danish respondents; (b) Inferred covariance matrix of the Danish respondents; (c-e) Individual moral principle values of three representative sub-sample of Danish respondents.}
    \label{fig:denmark}
\end{figure*}

We inferred the individual moral principles as well as the group values $w^g$ and the covariance matrices $\Sigma^g$. These results are shown in Figure \ref{fig:denmark}.
We note the variations in the inferred moral principles of three representative sub-sample of Danish respondents.

\subsection{Predicting Individual Judgments}
As an evaluation of our model, we performed out-of-sample prediction test. We randomly selected ten-thousand respondents from the Moral Machine website who completed at least one session, which contains thirteen scenarios. We filtered only the respondents' first 13 scenarios to compile a data set consisting 130,000 decisions.

We compared the predictive accuracy of the model against three benchmarks. Benchmark 1 models the collective values of the characters in Moral Machine such that the utility of a state is computed as  
\begin{equation}
u(\Theta) = w^{c \top} \Theta
\end{equation}
where $w^{c} \in \mathbb{R}^{K}$. Benchmark 1 models the weights as $w^{c} \sim Normal_{K}(\mu, \sigma^2 I)$ and does not include the group hierarchy or the covariance between the weights over the characters and factors (e.g. traffic light, passenger, etc.).     

Benchmark 2, which builds upon Benchmark 1, models the values along the abstracts moral dimensions $\Lambda$ as $w^{f} \sim Normal_{D}(\mu, \sigma^{2}I)$.  The group hierarchy and the covariance between weights are ignored.  

Finally, benchmark 3 models the individual moral principles of each respondent as $w^{l}_i \sim Normal_{D}(\mu, \sigma^{2}I)$, but does not include the hierarchical structure. Therefore, each respondent is viewed as an independent agent wherein inferring the values of one respondent provides no insight about the values of another.  

To demonstrate the gains in accuracy, we tested the models across different size of training data by varying the number of sampled respondents along $N=(4, 8, 16, 32, 64, 128)$. We used the first eight judgments from each respondent as training data, and tested the accuracy of predictions on the remaining five of the responses per each agent. For our model, we assumed that sampled respondents of size $N$ belong to one group.

\begin{figure}[h]
    \centering
    \includegraphics[width=0.4\textwidth]{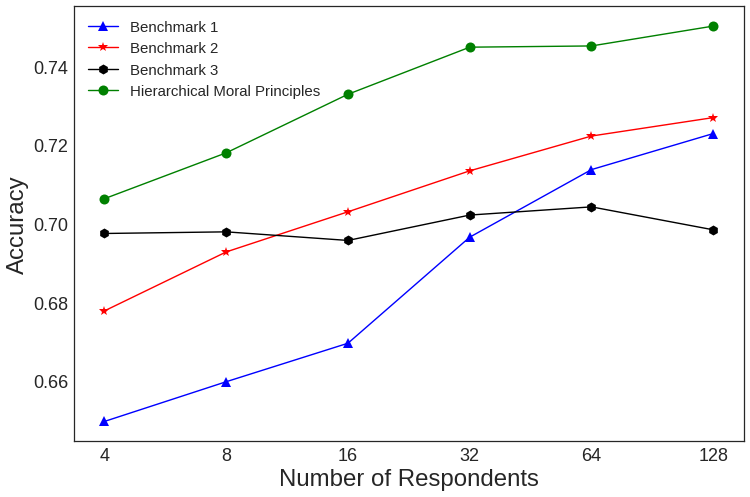}
    \caption{Comparison of out-of-sample prediction accuracy rates of the hierarchical moral principles model and the three benchmark models.}
    \label{fig:model_comparison}
\end{figure}

The results (Figure \ref{fig:model_comparison}) shows that as the number of respondents (i.e. training data) grows larger, predictive accuracy of our model, benchmark 1 and 2 improve. Accuracy of benchmark 3 does not improve as the the number of respondents have no bearing on inference of individual respondent's values. However, the hierarchical moral principles model shows consistently improving accuracy rates along the increasing size of the training data.  

We note that the margin of improvement between benchmark 1 and benchmark 2 reveals the gain achieved from abstraction and dimension reduction. The margin between benchmark 2 and our model reveals the gain from including individual moral principles. Finally, the margin between benchmark 3 and our model is indicative of the gain achieved by the group hierarchy.

\subsection{Response Time}
Studies in human decision making find strong relationship between the confidence level of the decision and reaction time of the decision (i.e. reaction time)  \citep{Smith2004PsychologyDecisions., Cain2012ComputationalNoise, Baron2017AJudgment}. These studies show that human subjects in binary-decision tasks take longer time to arrive at a decision when there is lower level of evidence. In this section, we take this approach to show that our model accurately captures the relationship between reaction time and difficulty of a moral dilemma.

We sampled 1727 respondents who accessed Moral Machine from the US; which altogether correspond to 22,451 judgments. In addition to the judgment decisions, we measured response times (RT) in seconds that the respondents took to arrive at their decisions.  Due to the unsupervised nature of the experiments, respondents are free to stop and reengage at later time; as such, we eliminated responses that took more than 120 seconds from our analysis. From the judgment data, after inferring the moral principles of individual respondents, we computed the estimated the probability of decision to swerve (i.e. $p^{t}_{i} = P(Y^{t}_{i} = 1|\Theta^{t}_{i})$) of each scenario as defined in Equation \ref{eq:sigmoid}. We computed new metric, {\em certainty of decision}, using $|p^t_{i} - 0.5|$.

\begin{figure}[h]
    \centering
    \includegraphics[width=0.4\textwidth]{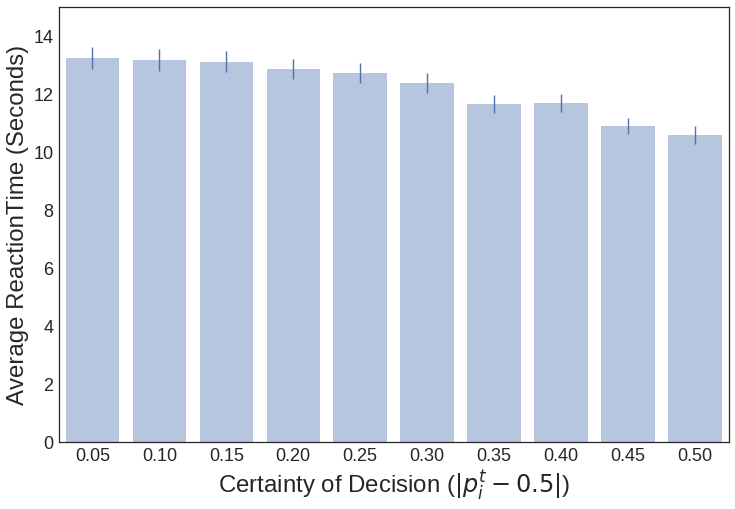}
    \caption{Reaction time in seconds per estimated certainty of decisions, which is defined as distance of the probability of judgment from the $0.5$ probability of swerving.}
    \label{fig:response_times}
\end{figure}

Plotting the certainty of decision and response times of the scenarios (see Figure \ref{fig:response_times}) reveals an intuitive pattern of relationship between two variables. 

Scenarios with higher certainty represent those that have clear trade-offs in the dilemmas such that the respondents on average respond quicker to the dilemmas.  Likewise, scenarios with lower certainties are those that have ambiguous trade-offs such that the respondents have less confidences about their decisions.  Intuitively, resolving the ambiguity of the trade-offs takes greater cognitive costs, which is revealed as longer response times for the respondents.

We view the relationship between response time and estimated certainty of decision from the model as a supporting evidence that the model is a robust representation of how people resolves moral dilemmas. In addition, the fact that cognitive cost of value-based decision process is revealed in their reaction times is an extra bit of information that could be used in the inference. For instance, we see a person making a quick decision; then we might also get information about the relative value difference between the two choices.  In future work, we intend to integrate response time information into the process of learning itself to allow the learner to infer even faster. 
 
%
%
\section{Discussion}
Drawing on a recent framework for modeling human moral decisions, we proposed a computational model of how the human mind arrives at a decision in moral dilemma.  We demonstrated the application of this model in the domain of autonomous vehicles using data from Moral Machine.  We showed that hierarchical Bayesian inference provides a powerful mechanism to accurately infer individual preferences as well as group norms along the abstract moral dimensions. We concluded with a demonstration of the model successfully capturing the cognitive cost in resolving the trade-offs in moral dilemmas. We show that moral dilemmas that are unpredictable by the model are correlated with long response times, where response times are a proxy of how difficult the dilemma is for the respondent because the subject is indifferent between the two responses. 

In this work, we have left out any discussion about method to aggregate the individual moral principles and the group norms to design an AI agent that makes decisions that optimize social utility of all other agents in the system. Recently, a paper by \cite{Noothigattu2017} introduced a novel method of aggregating individuals preferences such that the decision reached after the aggregation ensures global utility maximization. We view this method as a naturally complement to our work.   

Another interesting extension of our work is to explore the mechanism that maps the observable data on to the abstract feature space. We  formalized this process as feature mapping $F: \Theta \rightarrow \Lambda$.    Evidence from  developmental psychology suggests that children grow to acquire abstract knowledge and form inductive constraints \citep{Gopnik1997WordsTheories., Carey2009TheConcepts}. Non-parametric Bayesian processes such as the Indian Buffet Process \citep{ThomasL.Griffiths2005} and its variants \citep{Rai2009} are promising models to characterize this learning mechanism in the moral domain.

We used response time as a proxy to measure cognitive cost and proposed that the response time can be used as an extra information for more accurate inference over respondent's individual moral principles.  Combining our current model with drift diffusion model \citep{Ratcliff2008} can lead to a richer model that describes confidence and error in moral decision making.  An AI agent needs to understand moral basis of people's actions including when they are from socially inappropriate moral values as well as when they are mistakes. For instance, if an AI agent observes a person who spends a long time to make a ultimately wrong decision, the AI agent  should incorporate the person's confidence level and error rates to make accurate inference that the person most likely made a mistake.

Finally, we have inferred abstract moral principles and tested the model's predictive power on the same source of data. However, the abstract dimensions of the characters and factors in Moral Machine are not confined to the AV domain. An interesting experiment would be to test the model across various moral dilemmas in difference contexts. Hierarchical Bayesian models has been applied in the domain of transfer learning.  Demonstrating capacity to learn moral principles from one domain and apply these principles in other domains to make ethical decisions would show that a development of human-like ethical AI system does not need to be domain specific.

\newpage

\bibliography{mendeley}

\end{document}